\pdfoutput=1
\documentclass[11pt]{article}

\usepackage{acl}
\usepackage{tabularx}
\usepackage{geometry}
\usepackage{times}
\usepackage{latexsym}
\usepackage{amsmath}
\usepackage{amsfonts}
\usepackage{multirow}
\usepackage{booktabs,arydshln}
\usepackage{graphicx}
\usepackage{algorithm}
\usepackage{algorithmic}
\usepackage{tabu}
\usepackage{CJKutf8}
\usepackage{xcolor}
\usepackage{hyperref}
\usepackage{setspace}
\usepackage{latexsym}
\usepackage{amsmath}
\usepackage[T1]{fontenc}
\usepackage[utf8]{inputenc}

\usepackage{microtype}

\usepackage{inconsolata}

\usepackage{graphicx}

\title{Discovering Semantic Subdimensions through Disentangled Conceptual Representations}

\author{Yunhao Zhang$^{1,2}$, Shaonan Wang$^{1,2,}$\footnotemark[1], Nan Lin$^{3,4,}$\footnotemark[1], Xinyi Dong$^{5}$,  Chong Li$^{1,2}$, Chengqing Zong$^{1,2}$\\
  \footnotesize${}^1$State Key Laboratory of Multimodal Artificial Intelligence Systems, Institute of Automation, CAS, Beijing, China\\
  \footnotesize${}^2$School of Artificial Intelligence, University of Chinese Academy of Sciences, Beijing, China\\
  \footnotesize${}^3$State Key Laboratory of Cognitive Science and Mental Health, Institute of Psychology, CAS, Beijing, China\\
  \footnotesize${}^4$Department of Psychology, University of Chinese Academy of Sciences, Beijing, China\\
  \footnotesize${}^5$State Key Laboratory of Cognitive Neuroscience and Learning, Beijing Normal University\\
  \footnotesize{zhangyunhao2021@ia.ac.cn; shaonan.wang@polyu.edu.hk; linn@psych.ac.cn}
}

\begin{document}
\maketitle
\renewcommand{\thefootnote}{\fnsymbol{footnote}}
 %将脚注符号设置为fnsymbol类型，即特殊符号表示
% \footnotetext[2]{These authors contributed equally to this work.} %对应脚注[1]
\footnotetext[1]{Corresponding authors.}

\renewcommand{\thefootnote}{\arabic{footnote}}

\begin{abstract}
Understanding the core dimensions of conceptual semantics is fundamental to uncovering how meaning is organized in language and the brain. Existing approaches often rely on predefined semantic dimensions that offer only broad representations, overlooking finer conceptual distinctions. This paper proposes a novel framework to investigate the subdimensions underlying coarse-grained semantic dimensions. Specifically, we introduce a \textbf{D}isentangled \textbf{C}ontinuous \textbf{S}emantic \textbf{R}epresentation \textbf{M}odel (\textbf{DCSRM}) that decomposes word embeddings from large language models into multiple sub-embeddings, each encoding specific semantic information. Using these sub-embeddings, we identify a set of interpretable semantic subdimensions. To assess their neural plausibility, we apply voxel-wise encoding models to map these subdimensions to brain activation. Our work offers more fine-grained interpretable semantic subdimensions of conceptual meaning. Further analyses reveal that semantic dimensions are structured according to distinct principles, with polarity emerging as a key factor driving their decomposition into subdimensions. The neural correlates of the identified subdimensions support their cognitive and neuroscientific plausibility.

\end{abstract}

\section{Introduction}
Core dimensions are fundamental to structure mental representations, enabling systematic classification, context-sensitive interpretation, and generalization across novel situations \cite{allen1984towards,shepard1987toward,gardenfors2004conceptual}. In perceptual domains like vision, core dimensions include color, shape, motion, depth, and texture \cite{ge2022contributions,palmer1999vision,mapelli1997role}; in audition, they include pitch, loudness, timbre, spatial location, and rhythm \cite{poeppel2020speech,bizley2013and,temperley2004cognition}. These dimensions support object recognition (e.g., a red, round, shiny, motionless object = apple) and flexible inference (e.g., a loud sound in a forest = danger). 
% Importantly, by mapping stimuli onto a shared low-dimensional space, core dimensions allow perceptual systems to generalize from known examples to unfamiliar inputs. In contrast, conceptual semantics involves more abstract and less perceptually grounded dimensions, making their structure less transparent.
Crucially, this dimensional structure enables generalization by allowing novel stimuli to be interpreted based on their positions within a shared representational space.

% Core dimensions are fundamental to organizing mental representations of information, supporting systematic and flexible understanding \cite{allen1984towards,gardenfors2004conceptual,shepard1987toward}. In perceptual domains, such as action and face traits, these dimensions are relatively well-defined—for example, action perception is structured along dimensions like formidableness, friendliness, planned and abduction \cite{vinton2024four,barraclough2024perception}, while face traits involves trustworthiness, dominance, youthful attractiveness and sexual dimorphism \cite{oosterhof2008functional,sutherland2013social}. However, in the domain of conceptual semantics, the underlying dimensions are less transparent, as they encompass more abstract and non-perceptual aspects of meaning. 

Extending this framework to conceptual semantics is more challenging, as the underlying dimensions are more abstract and less perceptually grounded. 
One promising approach to addressing the above challenges is to empirically define a set of semantic dimensions \cite{binder2016toward,fernandino2016concept,diveica2023quantifying,wang2023large}. \citet{binder2016toward} introduced 65 semantic dimensions grounded in neuroscience research on conceptual representations. However, recent studies have shown that these 65 dimensions exhibit substantial overlap and lack neurobiological plausibility \cite{wang2022fmri,fernandino2022decoding,zhang2025simple}. To address this, \citet{wang2023large} proposed six major semantic dimensions: vision, action, space, time, social, and emotion. Follow-up neuroimaging studies have validated the robustness of these six dimensions \cite{zhang2025simple,tang2025neural,lin2024organization}. Nevertheless, this set of interpretable dimensions offers a coarse-grained representation of semantic space, which is limited in its ability to capture finer semantic distinctions that are crucial for accurately modeling the conceptual meaning \cite{binder2016toward, hoffman2013shapes}.

% This paper proposes an alternative framework to investigate the subdimensions underlying the six core semantic domains and their neural representational patterns.
This paper proposes a novel framework for investigating the subdimensions underlying six semantic dimensions. To establish neural validity, we examine their corresponding representational patterns in the brain. 
Specifically, we introduce a Disentangled Continuous Semantic Representation Model (DCSRM), which decomposes word embeddings from large language models into multiple semantic-specific sub-embeddings. Each sub-embedding is learned using a multi-objective optimization approach to maximize the encoding of its target semantic while minimizing interference from others.

We then interpret the meaning of the semantic subdimensions captured by these sub-embeddings by inspecting words with high and low loadings. To assess their neural plausibility, voxel-wise encoding models \cite{huth2016natural} are applied to map these subdimensions onto brain responses during natural language comprehension.

Our work offers a more fine-grained and interpretable set of semantic subdimensions for representing conceptual meaning than the conventional six semantic dimensions. These subdimensions are shown to be structured according to distinct principles, with polarity emerging as a key factor driving their decomposition. Moreover, we identify neural correlates of these subdimensions, aligning with prior neuroimaging studies \cite{lindquist2012brain, lin2024organization}, thereby supporting their cognitive and neuroscientific validity.

To summarize, our main contributions include:
\vspace{-0.5em}
\begin{itemize}
    \setlength{\itemsep}{0pt}
    \setlength{\parsep}{0pt}
    \setlength{\parskip}{0pt}
    \item We propose a disentangled continuous semantic representation model that separates distinct type of semantic information from large language models and extracts subdimensions within six semantic dimensions.

    \item We employ interpretable analysis method to define the meaning of each subdimension and identify key factors driving the decomposition of semantic subdimensions.

    \item We use neural encoding models to reveal the brain representations of the identified subdimensions, further supporting their cognitive plausibility.
\end{itemize}

\section{Related work}
\subsection{Conceptual semantic dimensions}
% Investigating the dimensions of conceptual semantics is crucial for understanding how meaning is organized in language, how humans use language to express, categorize, and reason about the world \cite{allen1984towards,gardenfors2004conceptual,shepard1987toward}. Moreover, semantic dimensions provide key insights into how word meaning is represented in the brain, helping to reveal the neural subsystems that support different types of semantic content and their organizational principles. 

Conceptual semantic dimensions are typically defined using either experience-based or data-driven approaches. Experience-based approaches, grounded in neuroscience and psychology, offer a solid foundation for defining semantic dimensions. \citet{binder2016toward} proposed 65 semantic dimensions, which have been shown to explain semantic-related behavior and brain activation \cite{anderson2017predicting,anderson2019integrated,tong2022distributed,fernandino2022decoding}. However, the 65 semantic dimensions exhibit notable overlap and redundancy. Specifically, some pairs of dimensions exhibit Pearson correlation coefficients exceeding 0.8, indicating a lack of independence between them \cite{wang2022fmri}. Furthermore, these dimensions do not all contribute equally to the explanation of brain activation. For instance, \citet{zhang2025simple}, using brain decoding methods, found that several non-sensorimotor dimensions (e.g., near, toward, away, number, benefit, needs) were not broadly represented across brain semantic networks. Building on these prior findings, \citet{wang2023large} proposed six coarse-grained semantic dimensions (vision, action, space, time, social, and emotion) and developed the Six Semantic Dimension Database, which contains subjective ratings for 17,940 Chinese words. Follow-up neuroimaging studies have further validated the effectiveness of these six dimensions \cite{zhang2023navigating,lin2024organization,zhang2025simple,tang2025neural}.

Data-driven approaches leverage the rich semantic information encoded in language models to uncover semantic dimensions by analyzing their word embeddings \cite{hollis2016principals,grand2022semantic}. \citet{hollis2016principals} applied PCA to skip-gram embeddings and identified emotion-related dimensions like valence and dominance. However, even after PCA, each dimension of embeddings often entangle multiple types of information. Moreover, skip-gram models capture less semantic richness than large language models (e.g., LLaMA, Alpaca), limiting their ability to identify fine-grained semantic dimensions.

% Combining the advantages of both approaches, we propose DCSRM, which uses semantic ratings from SSDD as labels and employs a multi-objective optimization framework to segment large language model representations into low-dimensional sub-embeddings, each capturing specific semantic information. We then analyze these sub-embeddings to identify semantic subdimensions.

\subsection{Disentangled methods}
Disentangled word representations serve two main purposes. First, they enhance interpretability by revealing the information encoded in embeddings \cite{karwa-singh-2025-disentangling,o2024disentangling,liao-etal-2020-explaining}; for instance, \citet{liao-etal-2020-explaining} decompose dense embeddings into sub-embeddings tied to discrete attributes (e.g., animal, location, adjective). Second, they benefit downstream NLP tasks such as sentence representation \cite{chen-etal-2019-multi}, text generation \cite{iyyer-etal-2018-adversarial}, word sense disambiguation \cite{silva-de-carvalho-etal-2023-learning}, and sentiment/style transfer \cite{zhu-etal-2024-styleflow}.
% Third, they help investigate the neural basis of language, with studies using fMRI and MEG to dissociate semantic and syntactic processing in the brain \cite{caucheteux2021disentangling,wang2020probing}.

Unlike prior work, this paper addresses a key question in psycholinguistics and neuroscience: What fine-grained semantic subdimensions enable more accurate representation of conceptual meaning, and to what extent are they grounded in neural activity? A central challenge lies in the overlapping and intertwined relationships among different semantic components within conceptual representations, making disentanglement inherently difficult. To address this, we propose a disentangled continuous semantic representation model (DCSRM), which segments large language model embeddings into multiple low-dimensional sub-embeddings, each encoding specific semantic information. We then analyze these sub-embeddings to identify semantic subdimensions and use voxel-wise encoding models to examine their neural correlates.

\section{Methods}
To investigate the subdimensions of each semantic dimension and their neural representations, we propose a three-step framework (Figure \ref{fig:frame}): 1) use DCSRM to encode each semantic information into a low-dimensional sub-embedding; 2) analyze the sub-embedding dimensions to identify semantic subdimensions; and 3) perform voxel-wise encoding to examine their relationship with brain activity during natural story comprehension.

\subsection{Disentangled continuous semantic representation model}
\label{sec:dcsrf}
To encode each semantic-specific information into low-dimensional sub-embeddings, we propose a disentangled continuous semantic representation model (DCSRM). Our goal is to transform $M$ $h$-dimensional dense word vectors $V \in \mathbb{R}^{M \times h}$ into disentangled embedding $X \in\mathbb{R}^{M\times h}$ by leveraging $N$ continuous semantic attributes $B = \{b_1, \dots, b_N\}$ labeled on words. 

$X$ is expected to have two properties. The first is retaining information encoded in $V$. More specifically, we require $V V^T \approx X X^T$ as pointed out by \citet{levy2014neural}, to ensure that the global similarity structure remains stable after the transformation. The second property is that the $h$ column vectors of $X$ are decomposed into $N{+}1$ sub-embedding sets, $X_{b_1}, \dots, X_{b_N}, X_{\text{unseen}}$, where each sub-embedding encodes information specific to one semantic attribute. For instance, $X_{b_1}$ is expected to represent information solely related to semantic attribute $b_1$, independent of the other semantic attributes $b_2, \dots, b_N$. To achieve these targets, we employ a multi-objective learning framework.

% The second property is that $X$的h个列向量经过N+1次下采样得到N+1个sub-embedding sets, $X_{b_1}, \dots, X_{b_N}, X_{unseen}$, where each sub-embedding encodes information specific to one semantic attribute.

\noindent\textbf{Orthogonal constraint ($\mathcal{L}_{\text{ORT}}$).} We transform the original embedding $V$ using a learnable projection matrix $W \in \mathbb{R}^{h \times h}$, yielding $X = V W$. To preserve the global semantic structure after transformation, we impose an orthogonality constraint $W^T W \approx I$, minimizing:
\begin{equation}
    \mathcal{L}_{\text{ORT}} = \| W^T W - I \|_2
\end{equation}
We have $X X^T = (VW)(VW)^T = V(WW^T)V^T = V V^T \quad \text{if} \quad WW^T = I \text{ holds.}$

\noindent\textbf{Continuous attribute prediction ($\mathcal{L}_{\text{SL}}$).} For modeling the relationship between sub-embeddings and continuous semantic attributes, we minimize the conditional expectation of prediction error:
\begin{equation}
    \mathbb{E}_{x_{b,j} \sim X_b, y_j \sim Y} \left[ | y_j - q_{\theta}(x_{b,j}) | \right]
\end{equation}
where $x_{b,j}$ denotes the $j$-th word representation in the sub-embedding matrix $X_b$ for attribute $b$, and $y_j$ is the corresponding ground-truth rating for that attribute. A parametric regression model $q_{\theta}(\cdot)$ is trained to predict $y_j$ from $x_{b,j}$ using the Smooth L1 loss:
\begin{equation}
    \mathcal{L}_{\text{SL}} = \sum_{j=1}^{N} \text{SmoothL1}(y_j, q_{\theta}(x_{b,j}))
\end{equation}
% where 
% \begin{equation*}
% \mathcal{L}_{\text{reg}} =
% \begin{cases}
% 0.5(y_j - q_{\theta}(x_{b,j}))^2 & \text{if } |y_j - q_{\theta}(x_{b,j})| < 1, \\
% |y_j - q_{\theta}(x_{b,j})| - 0.5 & \text{otherwise}.
% \end{cases}
% \end{equation*}

\noindent\textbf{Semantic contrastive loss ($\mathcal{L}_{\text{CE}}$).} To help the sub-embeddings capture more attribute-specific information, for each semantic attribute $b$, we treat words with ratings greater than a threshold as positive examples, and the others as negative examples. We minimize the contrastive cross-entropy loss:

\begin{equation}
    \mathcal{L}_{\text{CE}} = -\frac{1}{N} \sum_{i=1}^{N} \log \frac{\exp(\frac{x_i^\top x_i^{+}}{\tau})}{\sum_{j=1}^{N} \exp(\frac{x_i^\top x_j}{\tau})}
\end{equation}
where $x_i$ and $x_j$ denote the $i$-th and $j$-th words in the embedding space $X$, $x_i^{+}$ is the positive sample of $x_i$, and $\tau$ is a temperature parameter.

\noindent\textbf{Reconstruction loss ($\mathcal{L}_{\text{REC}}$).} To preserve input information while supporting semantic disentanglement, we let $X_b$ be features to reconstruct original vectors for words having attribute $b$ by minimize the reconstruction error:
\begin{equation}
    \mathcal{L}_{\text{REC}} =\| v_j - \varphi(x_{b,j}) \|^2
\end{equation}    
where $\varphi$ is a single fully connected layer mapping sub-embeddings back to the original space.

\noindent\textbf{KL-based sparsity constraint ($\mathcal{L}_{\text{KL}}$).} 
To promote feature selection and disentanglement, we apply variational dropout \cite{molchanov2017variational,liao-etal-2020-explaining} to each dimension of $X$. In the training process, we inject multiplicative noise on $X$:
\begin{equation}
  \xi \sim \mathcal{N} \left( 1, \alpha_b = \frac{p_b}{1 - p_b} \right)  
\end{equation}
where $p_b = sigmoid(\log \alpha_b)$ is the $h$-dimension dropout rates. For each attribute $b$ in $B$, the dimensions with dropout rates lower than 40\% are normally regarded as $X_b$.

% The $h$-dimensional dropout rates $p_b = sigmoid(\log \alpha_b)$ are then obtained. During each training iteration, a set of multiplicative noise $\xi$ is sampled from a normal distribution $\mathcal{N} \left( 1, \alpha_b = \frac{p_b}{1 - p_b} \right)$ and injected on $X$. We apply it during both prediction $\theta(\xi \odot X)$ and $\varphi(\xi \odot X)$. The $h$-dimensional dropout rates $p_b = sigmoid(\log \alpha_b)$ are then obtained. 

\noindent\textbf{Distribution alignment loss ($\mathcal{L}_{\text{DIS}}$).}
To encourage disentanglement when handling multiple attributes, we include a loss function on dropout rates. Let a $N$-dimensional vector $P$ be $1-p_b$ for all $b$ in $B$ in a specific dimension. The idea is to minimize $\prod_{j=1}^{N} p_j$ with constraint $\sum_{j=1}^{N} p_j = 1$. The optimal solution is that the dimension is relevant to only one attribute $b'$ where $1 - p_b' \approx 1$. In implementation, we minimize the following loss function:
\begin{equation}
    \mathcal{L}_{\text{DIS}} = \sum_{j=1}^{N} \log P_j + \beta \left\| \sum_{j=1}^{N} P_j - 1 \right\|^2
\end{equation}
We empirically set $\beta = 1$ following \citet{liao-etal-2020-explaining} to balance the trade-off between encouraging sparsity and maintaining normalized dimension weights, ensuring effective disentanglement.

\subsection{Sub-embedding analysis}
\label{sec:subemb}
To identify the semantic subdimensions encoded in each sub-embedding, we first apply Principal Component Analysis (PCA) to orthogonalize the dimensions within each sub-embedding, yielding transformed sub-embeddings $X'_{b_1}, X'_{b_2}, \dots, X'_{b_N}$. We then compute the Pearson correlation and pairwise order consistency between each dimension and the rating data, selecting those with significant correlations. For each, we instruct multiple large language models (e.g., GPT-4o, Grok3, Claude 3.7 Sonnet) as linguists to annotate the corresponding semantic subdimension based on the top-ranked words. Finally, we combine the results from these models to derive the final subdimension labels.

\subsection{Voxel-wise encoding models}
Voxel-wise encoding models map each transformed sub-embedding to its corresponding fMRI signal. To align with BOLD signal delay, word vectors are convolved with a canonical hemodynamic response function (HRF)\footnote{The canonical HRF models the expected BOLD response to a neural event.} and downsampled to match the fMRI sampling rate. During training, 14 additional regressors—capturing low-level stimulus properties such as word rate, word length, part-of-speech (adverb, noun, particle, verb), sound envelope, word frequency, and six head motions—are included but excluded during prediction.

We employ ridge regression and perform 5-fold nested cross-validation to ensure robust evaluation. Model performance is assessed by computing the Pearson correlation between predicted and observed fMRI signals. Group-level statistical significance is determined by comparing the estimated correlations to a null distribution of correlations derived from two independent Gaussian random vectors of equivalent length \cite{huth2016natural}.

Next, we explored the neural correlates of the semantic subdimensions within the brain regions significantly associated with the transformed sub-embeddings. We compute the average weight matrix by averaging the cross-validation weight matrices. To ensure consistent interpretability of weight direction, we apply sign correction to the weight matrix based on the correlation between each transformed sub-embedding dimension and rating data, as described in Section \ref{sec:subemb}. Finally, for each voxel, we assign the semantic subdimension with the highest weight as its representative subdimension.

% Creating the table
\setlength{\tabcolsep}{1.5mm}
\begin{table*}[ht]
	\centering
	\small
	\renewcommand\arraystretch{1.1}
	\vspace{-0.4cm}
    \resizebox{\textwidth}{!}{
\begin{tabular}{lcccccccccccccc}
\toprule
 & \multicolumn{2}{c}{$\text{DCSRM}_{vis}$} & \multicolumn{2}{c}{$\text{DCSRM}_{act}$} & \multicolumn{2}{c}{$\text{DCSRM}_{soc}$} & \multicolumn{2}{c}{$\text{DCSRM}_{emo}$} & \multicolumn{2}{c}{$\text{DCSRM}_{time}$} & \multicolumn{2}{c}{$\text{DCSRM}_{spc}$} & \multicolumn{2}{c}{Average} \\
\cmidrule(lr){2-3} \cmidrule(lr){4-5} \cmidrule(lr){6-7} \cmidrule(lr){8-9} \cmidrule(lr){10-11} \cmidrule(lr){12-13} \cmidrule(lr){14-15}
 & vis $\uparrow$ & non\_vis $\downarrow$ & act $\uparrow$ & non\_act $\downarrow$ & soc $\uparrow$ & non\_soc $\downarrow$ & emo $\uparrow$ & non\_emo $\downarrow$ & time $\uparrow$ & non\_time $\downarrow$ & spc $\uparrow$ & non\_spc $\downarrow$ & target $\uparrow$ & non\_target $\downarrow$ \\
\midrule
GloVe & 0.602 & 0.243 & 0.535 & 0.196 & 0.592 & 0.153 & 0.450 & 0.141 & 0.463 & 0.149 & 0.622 & \textbf{0.127} & 0.544 & 0.168 \\
Word2Vec & 0.826 & 0.228 & 0.702 & 0.245 & 0.792 & 0.189 & 0.654 & 0.149 & 0.649 & 0.225 & 0.790 & \underline{0.161} & 0.736 & 0.200 \\
\midrule
MacBERT-large & 0.840 & \textbf{0.172} & 0.728 & 0.190 & \textbf{0.858} & 0.127 & \textbf{0.772} & \textbf{0.114} & \textbf{0.807} & 0.160 & 0.855 & 0.213 & 0.810 & 0.163 \\
LLaMA2-1.3b & 0.881 & \underline{0.204} & 0.750 & 0.325 & 0.854 & \underline{0.058} & 0.770 & 0.118 & \underline{0.806} & \textbf{0.123} & 0.874 & 0.168 & 0.823 & 0.166 \\
Alpaca2-1.3b & 0.867 & 0.266 & 0.750 & 0.314 & 0.851 & 0.059 & \underline{0.771} & 0.116 & 0.796 & 0.136 & 0.861 & 0.174 & 0.816 & 0.177 \\
LLaMA2-7b & 0.886 & 0.219 & \underline{0.755} & 0.168 & \underline{0.856} & \textbf{0.056} & \textbf{0.772} & 0.121 & 0.804 & 0.130 & \underline{0.877} & 0.170 & \underline{0.825} & \textbf{0.144} \\
Alpaca2-7b & \underline{0.892} & 0.209 & \textbf{0.760} & \textbf{0.158} & 0.852 & 0.081 & \textbf{0.772} & 0.132 & \underline{0.806} & \underline{0.128} & \textbf{0.880} & 0.176 & \textbf{0.827} & \underline{0.147} \\
LLaMA3-8b & \textbf{0.896} & 0.223 & 0.730 & \underline{0.166} & 0.854 & \textbf{0.056} & 0.767 & \underline{0.115} & 0.771 & 0.148 & 0.851 & 0.172 & 0.812 & \underline{0.147} \\
\bottomrule
\end{tabular}
}
\caption{\textbf{Semantic prediction performance of DCSRM sub-embeddings.} For each semantic dimension (e.g., social), the “target” column (e.g., soc) reports the Pearson correlation between sub-embedding predictions and ground-truth ratings for the corresponding semantic dimension. The “non\_target” column (e.g., non\_soc) shows the average correlation with ratings from all other dimensions. Higher target and lower non\_target scores indicate better semantic disentanglement. Bolded values highlight the top-performing models, while underlined values indicate the second-best.}
 \vspace{-5mm}
        \label{tab1:prediction}
\end{table*}

\section{Experimental Setup}
\noindent\textbf{Brain imaging data.} We use the Chinese fMRI dataset from the SMN4Lang \cite{wang2022synchronized}, which was collected from 12 native speakers as they listened to 60 stories from the Renmin Daily Review\footnote{\url{https://www.ximalaya.com/toutiao/30917322/}}. These stories covered a broad range of topics, with each story lasting between 4 to 7 minutes, resulting in approximately 5 hours of audio content. The text and audio for all stories were downloaded from the Renmin Daily Review website, and the text was manually verified for consistency with the audio. The total word count across all stories was 43,326, forming a vocabulary of 9,153 unique words. After data collection, the fMRI data were preprocessed following the Human Connectome Project (HCP) pipeline \cite{glasser2013minimal}.

\noindent\textbf{Semantic rating data.} Our study utilized the rating dataset from the SSDD \cite{wang2023large}, which includes subjective ratings for 17,940 Chinese words. It focuses on six semantic dimensions: vision, action, social, emotion, space, and time. These 17,940 words encompass nearly all commonly used Chinese words, including verbs, nouns, adjectives, adverb, and quantifiers, etc. Table \ref{tab:definition} presents the definitions for each semantic dimension. Thirty human raters evaluated each word on a 1–7 scale (7 = very high, 1 = very low) for all six semantic dimensions, based on the given definitions. For each word on each dimension, the final rating was obtained by averaging the 30 individual scores. These semantic ratings accurately reflect the extent to which a concept involves information related to each dimension. Table \ref{tab:semantic_ratings} shows six-dimension semantic ratings for several concepts. For instance, the word "justice" received a score of 1 on the time dimension and 1.133 on the space dimension, indicating that it lacks temporal information and contains only a minimal amount of spatial information.

\noindent\textbf{Word representations.} We utilize eight widely adopted Chinese computational language models, categorized into two groups: context-independent models, including Word2Vec \cite{mikolov2013efficient} and GloVe \cite{pennington2014glove}, and context-aware models, comprising MacBERT-large \cite{cui-etal-2020-revisiting}, LLaMA2 (1.3B and 7B) \cite{touvron2023llama}, Alpaca2 (1.3B and 7B) \cite{taori2023stanford}, and LLaMA3 (8B) \cite{grattafiori2024llama}.

In line with prior research \cite{wang2024computational, zhang2023comprehensive, zhang2025mulcogbench}, for context-independent models, Word2Vec and GloVe embeddings were trained on the Xinhua News corpus (19.7 GB)\footnote{\url{http://www.xinhuanet.com/whxw.htm}} using identical model parameters. For context-aware models, we randomly sampled up to 1,000 sentences per target word from the Xinhua News corpus. These sentences were input into the models, and word vectors were extracted from the final layer. The vectors for each target word were then averaged to derive its word representation. To address substantial variations in hidden layer dimensions of context-aware models, we applied PCA to their word representations, reducing dimensionality while retaining at least 80\% explained variance.

\noindent\textbf{Evaluation.} We evaluate DCSRM using a semantic prediction task, where ridge regression predicts semantic ratings from sub-embeddings $X_b$. The Pearson correlation between predicted and ground-truth values is computed, and performance is assessed via five-fold nested cross-validation. To examine the impact of loss functions, we evaluate sub-embeddings obtained by removing each loss function individually. Additionally, we compare the performance of the original and disentangled embeddings to demonstrate that DCSRM retains the original information.

\setlength{\tabcolsep}{1.5mm}
\begin{table*}[ht]
	\centering
	\small
	\renewcommand\arraystretch{1.1}
	\vspace{-0.4cm}
    \resizebox{\textwidth}{!}{
\begin{tabular}{l *{14}{c}}
\toprule
 & \multicolumn{2}{c}{$\text{DCSRM}_{vis}$} & \multicolumn{2}{c}{$\text{DCSRM}_{act}$} & \multicolumn{2}{c}{$\text{DCSRM}_{soc}$} & \multicolumn{2}{c}{$\text{DCSRM}_{emo}$} & \multicolumn{2}{c}{$\text{DCSRM}_{time}$} & \multicolumn{2}{c}{$\text{DCSRM}_{spc}$} & \multicolumn{2}{c}{Average} \\
\cmidrule(lr){2-3} \cmidrule(lr){4-5} \cmidrule(lr){6-7} \cmidrule(lr){8-9} \cmidrule(lr){10-11} \cmidrule(lr){12-13} \cmidrule(lr){14-15}
 & vis $\uparrow$ & non\_vis $\downarrow$ & act $\uparrow$ & non\_act $\downarrow$ & soc $\uparrow$ & non\_soc $\downarrow$ & emo $\uparrow$ & non\_emo $\downarrow$ & time $\uparrow$ & non\_time $\downarrow$ & spc $\uparrow$ & non\_spc $\downarrow$ & target $\uparrow$ & non\_target $\downarrow$ \\
\midrule
DCSRM-$\mathcal{L}_{\text{DIS}}$ & \underline{0.879} & 0.820 & \textbf{0.770} & 0.842 & \textbf{0.863} & 0.824 & \textbf{0.794} & 0.837 & \textbf{0.805} & 0.835 & \textbf{0.870} & 0.822 & \textbf{0.830} & 0.830 \\
DCSRM-$\mathcal{L}_{\text{KL}}$ & 0.866 & \underline{0.207} & \underline{0.734} & 0.168 & \underline{0.854} & \underline{0.056} & 0.766 & \textbf{0.115} & 0.771 & \underline{0.144} & 0.849 & \underline{0.171} & 0.807 & \textbf{0.144} \\
DCSRM-$\mathcal{L}_{\text{SL}}$ & \textbackslash & \textbackslash & 0.461 & 0.572 & \textbackslash & \textbackslash & 0.162 & 0.171 & \textbackslash & \textbackslash & \textbackslash & \textbackslash & \textbackslash & \textbackslash \\
DCSRM-$\mathcal{L}_{\text{REC}}$ & 0.864 & 0.227 & 0.723 & \textbf{0.156} & 0.853 & \textbf{0.055} & 0.766 & \underline{0.133} & \underline{0.772} & \textbf{0.139} & 0.849 & \textbf{0.170} & 0.804 & \underline{0.147} \\
DCSRM-$\mathcal{L}_{\text{CE}}$ & 0.866 & \textbf{0.206} & 0.725 & \underline{0.160} & 0.851 & 0.060 & 0.762 & \textbf{0.115} & \underline{0.772} & 0.150 & 0.849 & 0.197 & 0.804 & 0.148 \\
DCSRM & \textbf{0.896} & 0.223 & 0.730 & 0.166 & \underline{0.854} & \underline{0.056} & \underline{0.767} & \textbf{0.115} & 0.771 & 0.148 & \underline{0.851} & 0.172 & \underline{0.812} & \underline{0.147} \\
\bottomrule
\end{tabular}
}
\caption{\textbf{Ablation study of DCSRM training on LLaMA3-8B.} The table reports semantic prediction performance across six semantic dimensions. For each semantic dimension (e.g., \textit{soc}), the “target” column reports the Pearson correlation between the predicted and ground-truth ratings for that dimension, based on its corresponding sub-embedding obtained without a specific loss (e.g., $\mathcal{L}_{\text{DIS}}$). The “non\_target” column reports the average correlation with ratings from all other dimensions. A slash (\textbackslash) indicates that no valid sub-embeddings are formed under the corresponding dropout setting. Higher target and lower non\_target scores indicate better semantic disentanglement. Bolded values denote the best results, and underlined values indicate the second-best. Loss terms: $\mathcal{L}_{\text{DIS}}$ = distribution alignment loss, $\mathcal{L}_{\text{KL}}$ = KL-based sparsity constraint, $\mathcal{L}_{\text{SL}}$ = continuous attribute prediction loss, $\mathcal{L}_{\text{ORT}}$ = orthogonality constraint, $\mathcal{L}_{\text{REC}}$ = reconstruction loss, and $\mathcal{L}_{\text{CE}}$ = semantic contrastive loss.}

	\label{tab2:xiaorong}
 \vspace{-5mm}
\end{table*}

\section{Results and Analysis}
\subsection{DCSRM result}
To assess the models’ capacity for semantic disentanglement, we report the prediction performance of DCSRM sub-embeddings across six semantic dimensions in Table~\ref{tab1:prediction}. As shown, DCSRM consistently produces well-disentangled representations across all language models: each sub-embedding captures rich, dimension-specific information while suppressing unrelated semantics. While Alpaca2-7B and LLaMA2-7B achieve the best average performance, the top-performing model differs by dimension. For instance, in the vision dimension, LLaMA3-8B better captures vision-specific semantics. We therefore select the best-performing model for each dimension in the subsequent subdimension analysis. 

Additionally, we observe that 7B models often produce sub-embeddings with richer target-specific semantic content than their 1.3B counterparts, and Alpaca2 achieves performance comparable to LLaMA2. This suggests that moderate parameter increases enhance semantic disentanglement, and that models retain fine-grained semantic capabilities even after supervised fine-tuning. Furthermore, context-aware models (e.g., LLaMA2, Alpaca2) consistently outperform context-independent models (e.g., Word2Vec, GloVe). This highlights their advantage in capturing dimension-specific semantic nuances and effectively suppressing irrelevant information, resulting in clearer semantic disentanglement.

Table \ref{tab2:xiaorong} reports the performance of LLaMA3-8b semantic-specific sub-embeddings on the semantic prediction task after ablating each loss function in DCSRM. The results show that removing either the $\mathcal{L}_{\text{DIS}}$ or the $\mathcal{L}_{\text{SL}}$ leads to the disappearance of sub-embeddings under certain dropout conditions or the emergence of entangled representations that mix target and non-target semantics. These findings indicate that $\mathcal{L}_{\text{SL}}$ (continuous attribute prediction loss) preserves magnitude differences within sub-embeddings, aligning attribute strengths with ground truth and facilitating effective subspace extraction via dropout. $\mathcal{L}_{\text{DIS}}$ (distribution alignment loss) promotes decorrelation among sub-embeddings, and its removal results in overlapping representations and weakened semantic boundaries. Moreover, the inclusion of $\mathcal{L}_{\text{KL}}$, $\mathcal{L}_{\text{REC}}$, and $\mathcal{L}_{\text{CE}}$ further improves the separation between target and non-target semantics within the sub-embedding space. 

We also observe that the original embeddings and the disentangled embeddings achieve comparable performance on the semantic prediction task, demonstrating that DCSRM preserves the information from the original vectors\footnote{See Table \ref{tab3:ori_trans} for details.}. 

Overall, these results demonstrate that DCSRM effectively separates different types of semantic information into distinct sub-embeddings. While not aiming for complete disentanglement, our goal is to maximize the separation of semantic dimensions within word representations.

\begin{table*}[ht]
\centering
\scriptsize
\renewcommand{\arraystretch}{1.2}
\begin{tabular}{llll}
\toprule
\textbf{Dimension} & \textbf{PC} & \textbf{Representative Words} & \textbf{Semantic subdimension} \\
\midrule
vision & PC1 & glasses, umbrella, parrot, tortoise, toaster, camellia, pepper, mask & Static vision \\
       & PC2 & hook, chisel, search, leap, drill, break in, knock, patch & Dynamic vision \\
\midrule
action & PC1 & cry, weep, shout, yell, tears, roar, laugh, shed tears & Outburst acts \\
       & PC2 & blink, cry, flicker, tremble, jump, shed tears, twinkle, gasp & Micro-movements \\
       & PC3 & push-up, dive, handstand, fall, capture, karate, retreat, registration & Forceful acts \\
       & PC4 & kneel, take down, issue, kneel on ground, sink, bury, order, play chess & Downward acts \\
       & PC5 & applaud, cheer, celebrate, run, ski, beg, swim, applause & Functional body acts \\
       & PC6 & bow, push-up, bend over, kowtow, handstand, squat, lower head, salute & Bending/ritual acts \\
\midrule
social & PC1 & duel, revolution, military, alliance, comrades, dispute, opponent, arbitration & Conflict \\
       & PC2 & pass, goal, submission, vote, assist, delivery, community, bidding & Collaboration \& exchange \\
\midrule
emotion & PC1 & obsession, love, affection, hate, disappointment, happiness, passion, addiction & Emotional load \\
        & PC2 & death, disease, corruption, murder, tragedy, grief, violation, suffering & Negative valence \\
        & PC3 & love, affection, romance, passion, care, devotion, fondness, attachment & Positive valence \\
\midrule
time    & PC1 & tomorrow, next year, recent, millennium, long-term, era, ancient, years & Temporal span \\
        & PC2 & Northern Wei, Eastern Jin, Western Jin, Jin Dynasty, military governor, prefect, poverty alleviation, Friday & Historical change \\
        & PC4 & anniversary, annual, Christmas, birthday, holiday, winter, same day, that day & Commemorative events \\
        & PC5 & Qing Dynasty, late Qing, Yuan Dynasty, Song Dynasty, Southern Song, early Qing, late Qing, Ming Dynasty & Dynastic eras \\
\midrule
space   & PC1 & railway, suburban, frigate, destroyer, southeast, northwest, capital, total area & Regional locations \\
        & PC2 & urban-rural, north-facing, nomadic, assembly hall, launch site, carrier rocket, rural, touring & Sites \& orientation \\
        & PC3 & epicenter, climbing, Arctic, cliff, bridge surface, canyon, plateau, outer space & Extreme spaces \\
\bottomrule
\end{tabular}
\caption{\textbf{Semantic subdimensions identified across six semantic dimensions.} For each PC, we present representative words selected from one end of the dimension—specifically, 8 out of the top 20 highest-loading words that strongly reflect the target semantic subdimension. The opposite end of the PC typically contains words unrelated to that subdimension. PCs with limited semantic relevance (Action PC7, Time PC3, and Social PC3) are excluded from this analysis (see Section~\ref{sec:sub_ana} for details).}
\vspace{-5mm}
\label{tab4:sub}
\end{table*}

\subsection{Sub-embedding analysis result}
\label{sec:sub_ana}
All principal components (PCs) showed significant correlations with their corresponding semantic ratings ($p < 0.05$), indicating relevance to their target domains\footnote{See Figure~\ref{fig:ssdd_dcsrm} for details.}. However, PCs with weak correlations (r < 0.1), accounting for 13.04\% of the total PCs, were considered to encode minimal semantic-relevant information and excluded from subsequent subdimension analyses.

% We can see from Figure~\ref{fig:ssdd_dcsrm} that all principal components (PCs) show significant correlations with their corresponding semantic ratings ($p < 0.05$), indicating that each PC captures information relevant to its semantic domain. However, some PCs show weak correlations (less than 0.1), encoding minimal domain-relevant information, and are thus excluded from the subsequent analysis of semantic subdimensions.

% Table~\ref{tab4:sub} summarizes the semantic subdimensions and representative words for each of the six semantic domains. In our analysis, we observed that for each PC within a given semantic domain, only one end typically carries strong semantic relevance, while the opposite end is largely unrelated to the targeted semantic content. Therefore, we focused on the semantically informative end of each PC for interpretation. From the top 30 highest-loading words, we selected 8 representative examples to illustrate each subdimension.

To interpret the meaning of each subdimension, we present representative words from each PC in Table~\ref{tab4:sub}. As shown, each semantic dimension is subdivided into two or more subdimensions, indicating the finer-grained organization within each semantic dimension. Some subdimensions align with prior research, such as the presence of negative valence and positive valence within the \textbf{emotion} dimension \cite{lindquist2012brain,russell2003core}. We also reveal previously unexplored semantic subdimensions, such as the dynastic eras and history change subdimensions in the \textbf{time} dimension. Our work provides a more detailed, fine-grained description of conceptual semantics compared to the conventional six semantic dimensions.

A cross-dimension comparison reveals that semantic dimensions vary in the number, polarity, and interpretability of their subdimensions. The \textbf{action} dimension exhibits the most diverse structure, comprising six subdimensions related to affective outburst, physical motion, and ritual behavior. In contrast, \textbf{vision}, \textbf{social}, and \textbf{emotion} show more well-defined relationships between orthogonal subdimensions within their respective semantic dimensions, which are structured along clear semantic axes that indicate polarized relationships (e.g., static vs. dynamic vision, conflict vs. collaboration, positive vs. negative valence). This suggests that polarization serves as an important factor driving the decomposition of semantic dimensions into finer-grained subdimensions.

We also find that individual subdimensions span multiple semantic dimensions. For instance, the dynamic vision subdimension in the \textbf{vision} dimension encodes action-related information; some high-loading words in subdimensions of the \textbf{space} dimension also encode aspects of visual information. These findings align with prior research that suggests the original rating data for vision, action, and space dimensions exhibit relatively high correlations \cite{wang2023large,lin2024organization}.

Moreover, the data-driven decomposition reveals semantic structures that deviate from intuitive or traditional expectations. For instance, the \textbf{time} blends abstract temporal spans (e.g., millennium, recent) with culturally grounded markers (e.g., Qing Dynasty, Christmas), while the \textbf{space} ranges from concrete geographical references (e.g., suburban, north-facing) to symbolic or extreme locations (e.g., outer space, cliff).
These findings suggest that time and space may not be structured solely along geometric or chronological lines, but are enriched by cultural and experiential semantics. This may also reflect the nature of large language models trained purely on text, which tend to capture more abstract and culturally embedded semantic patterns.

Overall, these patterns reveal that conceptual semantic dimensions are not flat or homogeneous, but instead exhibit structured subspaces organized along psychologically meaningful dimensions. This supports the view that human semantic knowledge is shaped by a small set of interpretable principles, such as polarity and hierarchy \cite{rosch1975cognitive,lakoff2008women,gardenfors2004conceptual}.

% The \textbf{vision} space revealed a concreteness axis (e.g., glasses, umbrella, parrot) and a dynamism axis (hook, chisel). Within the \textbf{action} space, we identified six subdimensions: emotional outbursts (cry, shout), small-scale bodily movements (blink, tremble), forceful movements (push-up, capture), downward-oriented acts (kneel, sink), social interactions (applaud, cheer), and ritualized postures (bow, salute). The \textbf{social} dimension encoded axes of conflict (duel, revolution) and cooperative exchange (vote, community). The \textbf{emotion} space decomposed into emotional intensity (love, obsession), negative valence (death, corruption), and positive valence (care, attachment), offering a fine-grained affective structure. For \textbf{time}, principal components reflected temporal span (tomorrow, millennium), sociohistorical transition (Northern Wei, poverty alleviation), commemorative events (birthday, Christmas), and dynastic eras (Qing Dynasty, Southern Song), highlighting both abstract and culturally grounded temporal concepts. Finally, the \textbf{space} dimension revealed substructures related to macro-geographical regions (southeast, capital), spatial orientation (north-facing, urban-rural), and boundary environments (cliff, outer space). 

\begin{figure*}[htbp]
\centering 
\setlength{\textfloatsep}{2pt}
\setlength{\intextsep}{2pt}
\setlength{\abovecaptionskip}{2pt}
\includegraphics[width=0.95\textwidth]{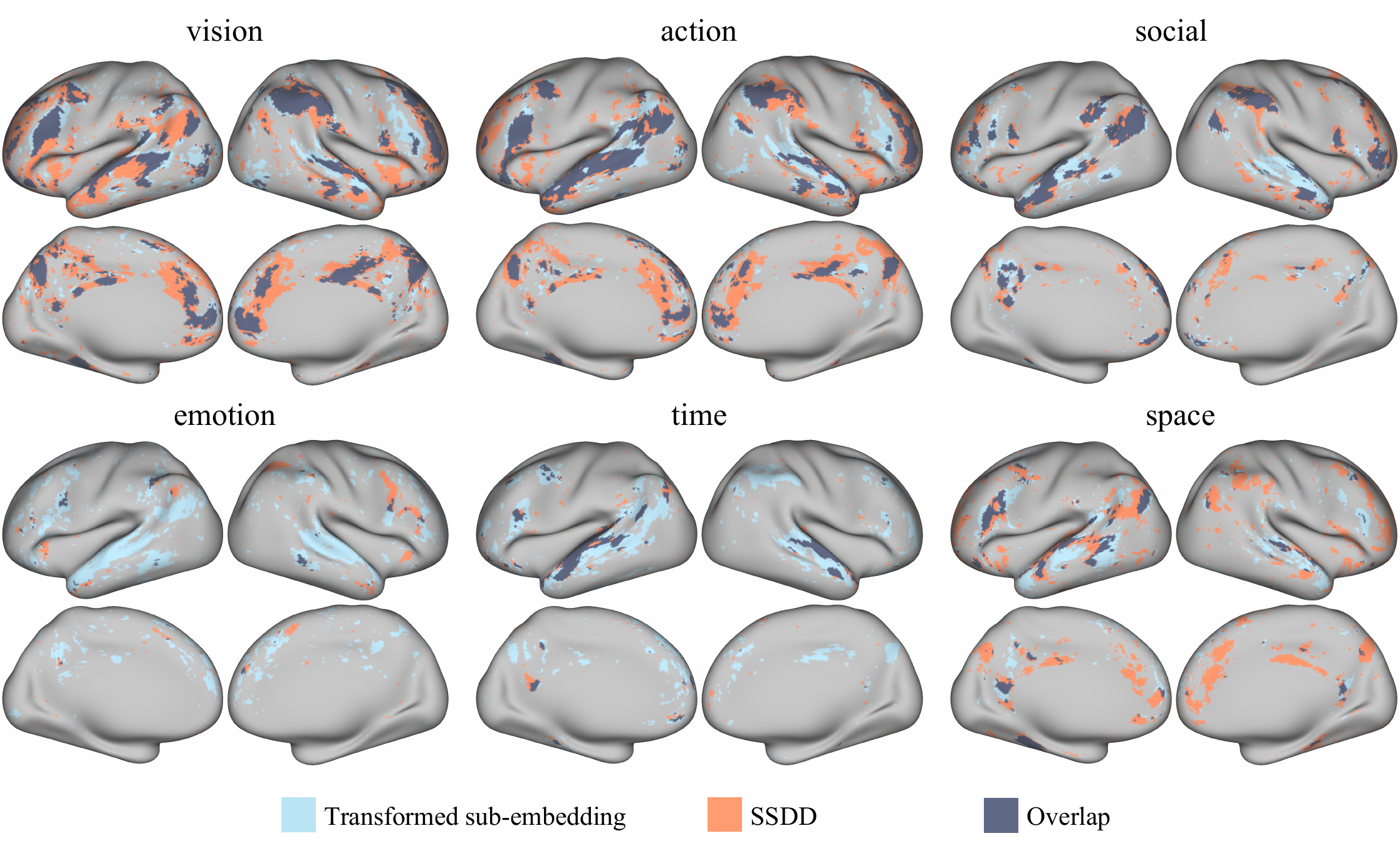}
\caption{\textbf{Distinct and overlapping neural correlates of SSDD ratings and transformed sub-embeddings across semantic dimensions.}
Blue regions indicate voxels where BOLD responses are significantly predicted by transformed sub-embeddings, while orange regions show the same for SSDD rating data (one-tailed t-test, $-\log(p)$, $p < 0.001$). Slate-gray regions denote overlap between the two. Transformed sub-embeddings denote the result of applying PCA separately to each semantic-specific sub-embedding generated by DCSRM (see Section~\ref{sec:subemb} for details). This figure integrates results from Figure~\ref{fig:ssdd_encoding} and Figure~\ref{fig:pca_encoding}.}
\label{fig:encoding}
\vspace{-5mm}
\end{figure*}

\subsection{Voxel-wise encoding result}
Figure \ref{fig:encoding} illustrates both distinct and overlapping neural correlates of SSDD ratings and transformed sub-embeddings. In embodied-related dimensions (e.g., vision, action), SSDD rating data predicts a broader range of brain regions compared to the transformed sub-embeddings. However, in abstract dimensions (e.g., emotion and time), the transformed sub-embeddings predict a wider array of neural activation patterns than the SSDD rating data. These findings suggest that languag language models trained on vast amounts of pure text data capture human-level conceptual representation in non-sensorimotor dimensions, but they capture relatively less sensorimotor information, consistent with recent studies \cite{xu2025large}. Moreover, the sub-embeddings activate brain regions including the inferior frontal gyrus (IFG), superior temporal gyrus (STG), posterior superior temporal sulcus (pSTS), middle temporal gyrus (MTG), inferior temporal gyrus (ITG), precuneus (Pcun), cingulate gyrus (CG), fusiform gyrus (FG), and angular gyrus (AG). These regions are largely consistent with prior findings on language-processing networks in the brain\footnote{See Figure~\ref{fig:BNA} for details.} \cite{binder2009semantic,huth2016natural,yang2019uncovering}, thereby supporting the validity of our computational framework for studying neural language comprehension. We also find that brain regions activated by different sub-embeddings has substantial overlap. A possible explanation is the multi-functionality of brain regions, where the same region represent multiple semantic subdimensions.

% Moreover, the sub-embeddings activate brain regions with substantial overlap, covering a significant portion of the semantic network \cite{binder2009semantic,huth2016natural,yang2019uncovering}, including the inferior frontal gyrus (IFG), superior temporal gyrus (STG), posterior superior temporal sulcus (pSTS), middle temporal gyrus (MTG), inferior temporal gyrus (ITG), precuneus (Pcun), cingulate gyrus (CG), fusiform gyrus (FG), and angular gyrus (AG). A possible explanation for the substantial overlap is the multifunctionality of brain regions, where the same region may represent multiple semantic subdimensions.

% Another reason is that some subdimensions may reflect interactions between semantic domains. For instance, two social subdimensions (conflict, collaboration \& exchange) represent the interaction between social and emotional domains. Overall, these findings also support the validity of our computational framework for studying neural language comprehension.

\begin{figure*}[ht]
\centering 
\setlength{\textfloatsep}{2pt}
\setlength{\intextsep}{2pt}
\setlength{\abovecaptionskip}{2pt}
\includegraphics[width=0.95\textwidth]{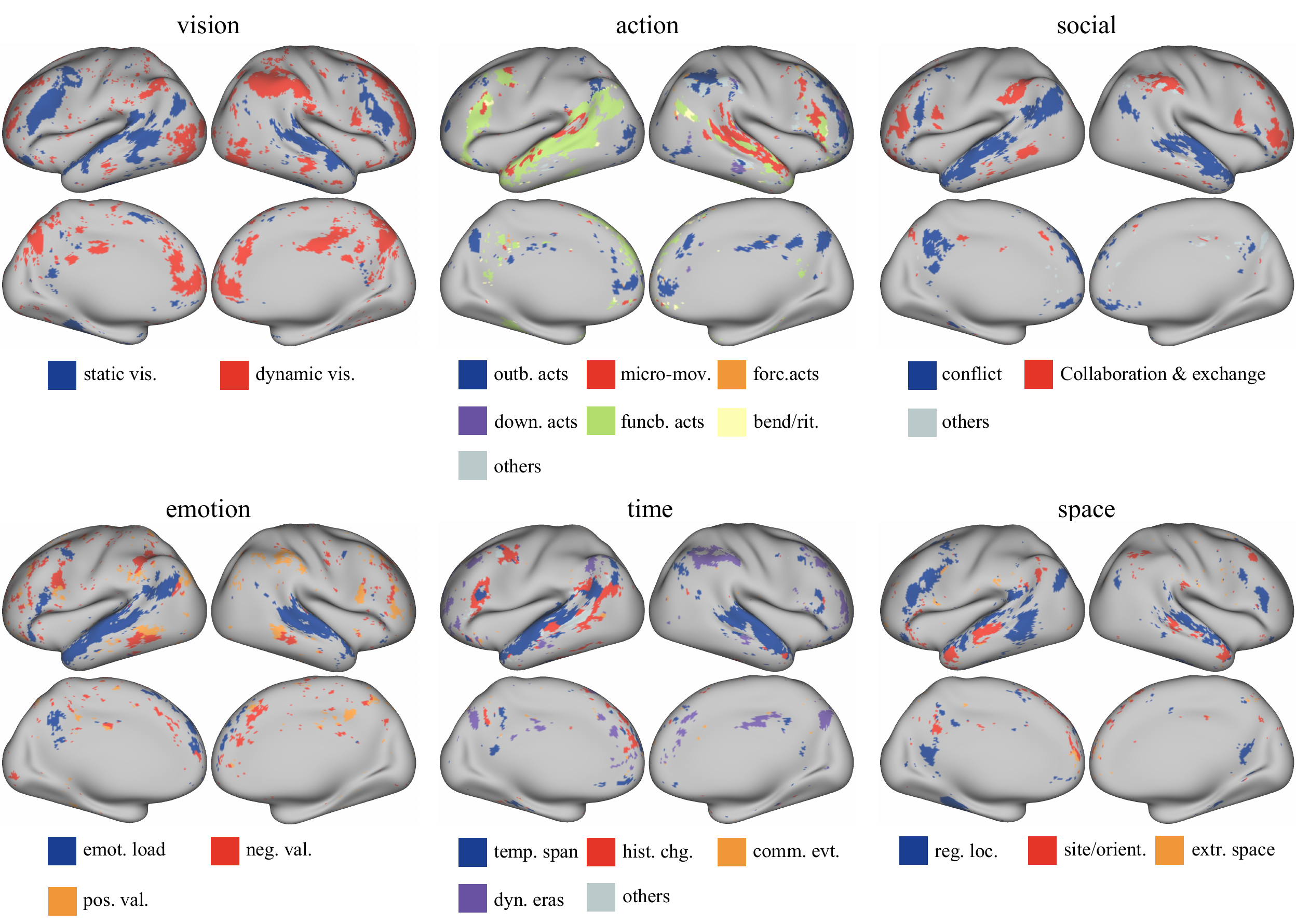}
\caption{\textbf{Neural correlates of semantic subdimensions across six semantic dimensions.} Others refers to subdimensions from the transformed sub-embeddings that carry minimal information related to the corresponding semantic dimension (see Section~\ref{sec:sub_ana} for details).}
\label{fig:subdimension}
\vspace{-5mm}
\end{figure*}

Figure \ref{fig:subdimension} shows neural correlates of semantic subdimensions. As shown, semantic subdimensions are distributively represented across semantic networks, indicating that these subdimensions rely heavily on distributed brain networks rather than localized brain regions. This finding aligns with the widely held view in cognitive neuroscience that the brain employs distributed representations to encode all types of information even primitive features \cite{zhang2022probing,lin2024organization}.  

Several subdimensions across different dimensions converge on similar brain regions, particularly the CG, IFG, MTG, and AG. For instance, collaboration \& exchange (social), positive valence (emotion), and dynastic eras (time) all activate regions involved in autobiographical memory, value processing, and social reasoning \cite{d2014brains,singer2001new,lin2020dissociating}. This suggests shared cognitive mechanisms underlying semantically rich, socially meaningful concepts. Moreover, we find that the left AG and left STG are involved not only in representing sensorimotor information (e.g., static vision, micro-movements) but also in encoding higher-level social and non-sensorimotor information (historical change, conflict, emotional valence). This supports a popular research view that the left AG and STG function as semantic hubs, integrating associations between sensory and social knowledge \cite{lin2018fine,skipper2011sensory}.

Our results are also consistent with prior neuroimaging studies. We find that the Pcun, right MTG, and right AG are involved in representing collaboration \& exchange, corroborating previous findings on the neural representation of collaborative behavior \cite{xie2020finding}. Brain regions associated with positive and negative valence also align with prior research on emotional processing \cite{kragel2016decoding,wager2015bayesian,saarimaki2016discrete}. The subdimension of static vision engages the AG, Pcun, posterior CG, and FG, consistent with known visual-semantic areas \cite{lin2018fine,sabsevitz2005modulation}. 

Beyond these, we also identify brain regions associated with certain semantic subdimensions that, to our knowledge, have not been reported in previous work. For instance, the dynastic eras activates the Pcun, MTG, and CG. The historical change activates the dorsal IFG, ITG, and AG. For Action, the micro-movements activates the right STG and right IFG, while functional body acts engage the left STG, left MTG, left IFG, and left Pcun. These results extend prior semantic neuroscience by identifying novel neural correlates for semantic subdimensions, advancing our understanding of how fine-grained conceptual semantics are encoded in the brain. Future work will further validate the reliability of these findings and explore their cognitive and neural significance in greater depth.

% Taken together, these findings demonstrate that semantic subdimensions are widely distributed across the brain, suggesting that fine-grained conceptual information is encoded through distributed neural representations.

\section{Conclusions}
Our primary goal is to investigate the fine-grained structure of conceptual semantics. A key challenge lies in defining and quantifying semantic subdimensions. To address this, we propose a Semantic Representation Disentangling Model with an interpretable framework that decomposes word embeddings into multiple sub-embeddings, each capturing distinct semantic content and corresponding to a specific subdimension. These subdimensions are shown to be structured according to distinct principles, with polarity emerging as a key factor driving their decomposition. Neural encoding analyses further confirm their representation in the brain. Compared to traditional approaches, our method automatically uncovers more fine-grained semantic dimensions from large-scale data without manually specifying their number or type, offering a data-driven foundation for constructing a conceptual semantic representation system that supports systematic classification, context-sensitive interpretation, and generalization to novel situations.

\section*{Limitations}
First, our analysis currently focuses only on the six semantic dimensions included in the SSDD database. Other dimensions, such as auditory and tactile, may also play a significant role in conceptual semantics. In future work, when large-scale rating data for additional semantic dimensions become available, we will expand the fine-grained semantic space we have developed and explore the neural correlates of a broader range of semantic dimensions in natural language understanding.

Second, we rely solely on large language models trained on text data to explore the subdimensions of each semantic dimension. However, human conceptual knowledge is acquired not only through abstract symbolic interactions but also through sensory and motor experiences \cite{bi2021dual,paivio1990mental}. Recent advances in multimodal models, which integrate both linguistic and sensory inputs, have demonstrated that these models more closely resemble human cognitive representations \cite{tang2021cortical,wang2023better}. In future work, we plan to integrate multimodal large models to further explore the fine-grained semantic space, enabling a more comprehensive investigation of semantic subdimensions.

Third, our findings are based on Chinese data, and it remains unclear whether these results can be generalized to other languages, such as English or German. In future studies, we plan to use large-scale rating data from other languages to further validate the stability of our conclusions.

Finally, our current work primarily utilizes the semantic subdimension framework to explore the neural representation patterns of fine-grained conceptual semantics. In future studies, we plan to investigate the broader potential of this framework in other domains, such as brain decoding \cite{sun2023contrast,ye2025decodingmultimodalmindgeneralizable}, model interpretability \cite{li-etal-2023-interpreting, duan2025syntax}, enhanced performance on semantically related tasks \cite{jian2025lookagainthinkslowly, chen2023chinesewebtext}, and the diagnosis and auxiliary treatment of neurological disorders \cite{dong2024disrupted, wang2025bridging}.

\section*{Ethical Statements}
We used preprocessed data from a publicly available dataset. All cognitive data were anonymized to ensure that no personally identifiable information was retained. The dataset was provided by the Institute of Automation, Chinese Academy of Sciences. All experimental procedures were approved by the Institutional Ethics Committee of the Institute of Psychology, Chinese Academy of Sciences, and the Institutional Review Board of Peking University, in accordance with established ethical guidelines and regulations.

\section*{Acknowledgements}
We would like to thank the anonymous reviewers for their helpful discussions and valuable comments. This research was supported by grants from the National Natural Science Foundation of China (No. 62036001) and the STI2030-Major Project (No. 2021ZD0204105).

% Bibliography entries for the entire Anthology, followed by custom entries
%\bibliography{anthology,custom}
% Custom bibliography entries only
% \bibliography{anthology,custom}
% \bibliographystyle{acl_natbib}

% \bibliography{references,anthology}
\appendix
\newpage

\renewcommand{\thetable}{A\arabic{table}}
\renewcommand{\thefigure}{A\arabic{figure}}

\section{Overview of the framework}
\begin{figure*}[htbp]
\centering 
\includegraphics[width=0.95\textwidth]{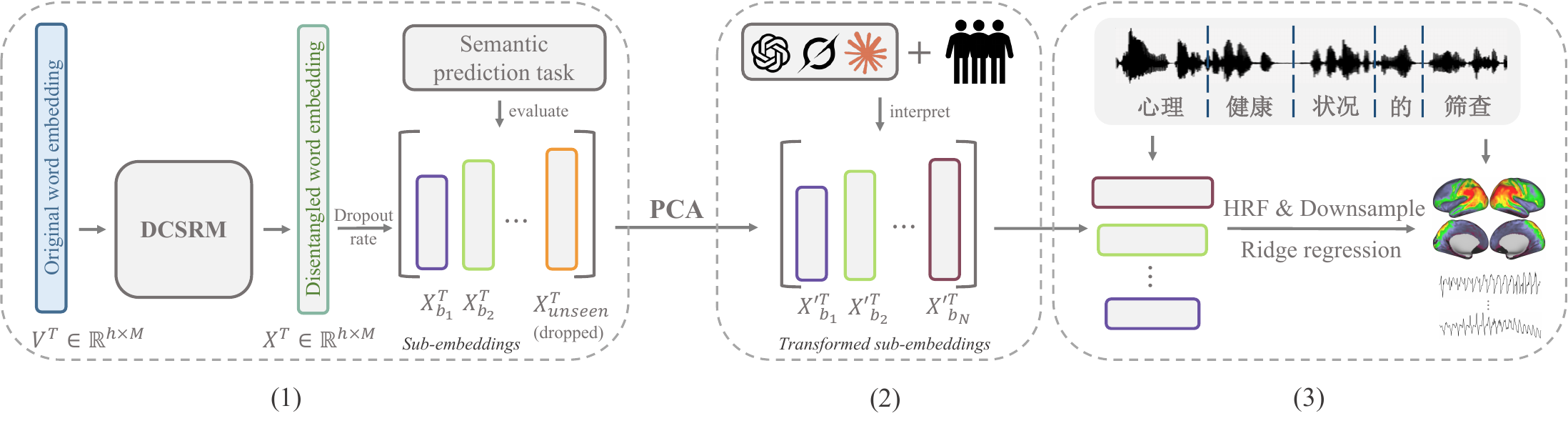}
\caption{\textbf{Overview of the proposed framework.} The framework consists of three key stages:
(1) Disentangled continuous semantic representation model (DCSRM). Original word embeddings $V \in \mathbb{R}^{M \times h}$ extracted from language models are input to DCSRM, producing disentangled embeddings $X \in \mathbb{R}^{M \times h}$ and a dropout rate matrix $D \in \mathbb{R}^{N \times h}$, where $M$ is the number of words, $h$ is the embedding dimension, and $N$ is the number of semantic attributes. Each element $(i,j)$ in $D$ indicates the selectivity of the $j$-th dimension of $X$ for the $i$-th attribute, with lower dropout rates denoting higher semantic relevance. Thresholding $D$ yields multiple attribute-specific sub-embeddings.
(2) Sub-embedding analysis. Each attribute-specific sub-embedding $X_{b_1}, X_{b_2}, \dots, X_{b_N}$ is orthogonalized via PCA to obtain transformed sub-embeddings $X'_{b_1}, X'_{b_2}, \dots, X'_{b_N}$. For each dimension within each transformed sub-embedding, we prompt multiple large language models (e.g., GPT-4o, Grok-3, Claude 3.7 Sonnet) to infer the underlying semantic meaning based on the top-ranked words. Their responses are aggregated to determine the final subdimension labels.
(3) Voxel-wise encoding. We map each transformed sub-embedding to brain activity during natural story comprehension using voxel-wise encoding, resulting in a weight matrix $U \in \mathbb{R}^{f \times g}$, where $f$ is the number of subdimensions and $g$ is the number of cortical voxels. Each voxel is assigned the subdimension with the highest weight, indicating its strongest semantic preference. The example stimulus "\begin{CJK*}{UTF8}{gbsn}心理健康状况的筛查\end{CJK*}" is taken from the fMRI dataset used in this study, and its English translation is “Screening for mental health conditions”.}
\label{fig:frame}
% \vspace{-5mm}
\end{figure*}

Figure \ref{fig:frame} shows the overview of the framework to identify the subdimensions within each semantic dimension, and use the voxel-wise encoding method to assess their neural plausibility.

\begin{table*}[htbp]
\centering
\scriptsize
\renewcommand{\arraystretch}{1.2}
\begin{tabular}{ll}
\toprule
\textbf{Dimension} & \textbf{Definition} \\
\midrule
Vision & The extent to which the meaning of a word can easily and quickly trigger corresponding visual images in your mind \\
Action & The extent to which the meaning of a word can easily and quickly trigger corresponding body actions in your mind \\
Social & The extent to which the meaning of a word relates to relationships or interactions between people \\
Emotion & The extent to which the meaning of a word relates to positive or negative emotions \\
Time & The extent to which the meaning of a word relates to time, including early or late, length, sequence, frequency, etc. \\
Space & The extent to which the meaning of a word relates to spatial information, including location, direction, distance, path, scene, etc. \\
\bottomrule
\end{tabular}
\caption{Definition of each semantic dimension.}
% \vspace{-5mm}
\label{tab:definition}
\end{table*}

\begin{table*}[hbp]
\centering
\scriptsize
\renewcommand{\arraystretch}{1.2}
\begin{tabular}{lcccccc}
\toprule
\textbf{Word} & \textbf{Vision} & \textbf{Action} & \textbf{Social} & \textbf{Emotion} & \textbf{Time} & \textbf{Space} \\
\midrule
justice & 1.600 & 1.667 & 2.400 & 3.867 & 1.000 & 1.133 \\
sea & 6.033 & 2.767 & 1.133 & 1.500 & 1.000 & 5.933 \\
june & 2.333 & 2.033 & 1.067 & 1.167 & 6.733 & 1.267 \\
football & 6.500 & 5.433 & 4.133 & 1.533 & 1.300 & 3.200 \\
of & 1.067 & 1.000 & 1.100 & 1.000 & 1.033 & 1.033 \\
\bottomrule
\end{tabular}
\caption{Six-dimension semantic ratings for the concepts "justice (\begin{CJK*}{UTF8}{gbsn}正义\end{CJK*})", "sea (\begin{CJK*}{UTF8}{gbsn}海洋\end{CJK*})", "june (\begin{CJK*}{UTF8}{gbsn}六月\end{CJK*})", "football (\begin{CJK*}{UTF8}{gbsn}足球\end{CJK*})" and "of (\begin{CJK*}{UTF8}{gbsn}的\end{CJK*})".}
% \vspace{-5mm}
\label{tab:semantic_ratings}
\end{table*}

\section{Validation result}
\setlength{\tabcolsep}{1.5mm}
\begin{table*}[t]
	\centering
	\small
	\renewcommand\arraystretch{1.1}
	\vspace{-0.4cm}
    \resizebox{\textwidth}{!}{
\begin{tabular}{lcccccccccccccc}
\toprule
& \multicolumn{2}{c}{vis} & \multicolumn{2}{c}{act} & \multicolumn{2}{c}{soc} & \multicolumn{2}{c}{emo} & \multicolumn{2}{c}{time} & \multicolumn{2}{c}{spc} & \multicolumn{2}{c}{Average} \\
\cmidrule(lr){2-3} \cmidrule(lr){4-5} \cmidrule(lr){6-7} \cmidrule(lr){8-9} \cmidrule(lr){10-11} \cmidrule(lr){12-13} \cmidrule(lr){14-15}
& origin & disent. & origin & disent. & origin & disent. & origin & disent. & origin & disent. & origin & disent. & origin & disent. \\
\midrule
GloVe        & 0.647 & 0.646 & 0.604 & 0.602 & 0.651 & 0.650 & 0.561 & 0.559 & 0.534 & 0.531 & 0.681 & 0.679 & 0.613 & 0.611 \\
Word2Vec     & 0.854 & 0.853 & 0.758 & 0.758 & 0.823 & 0.822 & 0.746 & 0.746 & 0.697 & 0.697 & 0.822 & 0.822 & 0.783 & 0.783 \\
\midrule
MACBERT      & 0.878 & 0.878 & 0.795 & 0.795 & 0.875 & 0.875 & 0.836 & 0.836 & 0.843 & 0.843 & 0.885 & 0.885 & 0.852 & 0.852 \\
LLaMA2-1.3b  & 0.906 & 0.900 & 0.835 & 0.831 & 0.885 & 0.885 & 0.849 & 0.857 & 0.876 & 0.872 & 0.900 & 0.898 & 0.875 & 0.874 \\
Alpaca2-1.3b & 0.893 & 0.893 & 0.816 & 0.816 & 0.879 & 0.879 & 0.842 & 0.842 & 0.869 & 0.869 & 0.894 & 0.894 & 0.865 & 0.865 \\
LLaMA2-7b    & 0.905 & 0.904 & 0.831 & 0.831 & 0.884 & 0.884 & 0.850 & 0.850 & 0.876 & 0.877 & 0.900 & 0.900 & 0.874 & 0.874 \\
Alpaca2-7b   & 0.906 & 0.906 & 0.835 & 0.835 & 0.885 & 0.884 & 0.849 & 0.849 & 0.876 & 0.876 & 0.900 & 0.900 & 0.875 & 0.875 \\
LLaMA3-8b    & 0.908 & 0.908 & 0.799 & 0.799 & 0.874 & 0.874 & 0.834 & 0.834 & 0.857 & 0.858 & 0.885 & 0.885 & 0.860 & 0.860 \\
\bottomrule
\end{tabular}
}
\caption{\textbf{Semantic prediction accuracies on original and disentangled embeddings.} For each semantic dimension (e.g., vision), columns labeled “origin” and “disent.” report the Pearson correlation between predicted and ground-truth ratings using original and disentangled embeddings, respectively. “Origin” refers to the original embeddings $V \in \mathbb{R}^{M \times h}$ extracted from language models, and “disent.” refers to the disentangled embeddings $X \in \mathbb{R}^{M \times h}$ obtained by inputting $V$ into the DCSRM model, where $M$ is the number of words and $h$ is the embedding dimension.}
	\label{tab3:ori_trans}
 % \vspace{-5mm}
\end{table*}

Table \ref{tab3:ori_trans} shows the semantic prediction accuracies on the original and disentangled embeddings.

\section{Semantic alignment of transformed sub-embddings.}

\begin{figure*}[htbp]
\centering 
\includegraphics[width=0.95\textwidth]{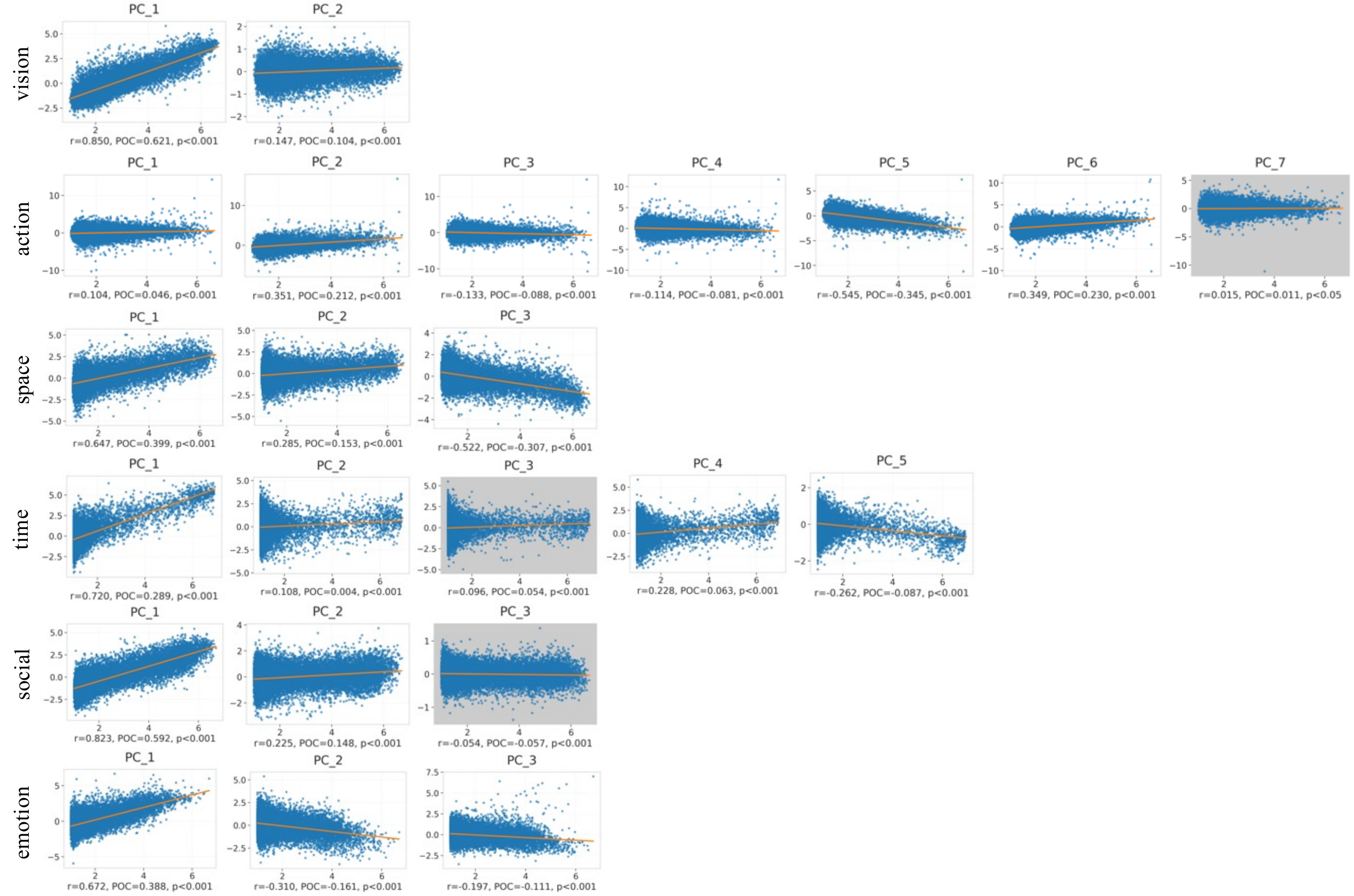}
\caption{\textbf{Semantic alignment of transformed sub-embddings.} Pearson correlation (r) and pairwise order consistency (POC) between the semantic rating data (x-axis) and each dimension of the transformsub-embedding within the corresponding semantic dimension (y-axis). We only annotate $p < 0.001$ on the figure when both correlation results are statistically significant at $p < 0.001$. Significant correlations ($p < 0.001$) indicate that the respective sub-embedding dimension encodes the corresponding semantic information. Dimensions with a correlation below 0.1 are shown with a gray background. Transformed sub-embeddings refer to the outputs obtained by applying PCA individually to each semantic-specific sub-embedding produced by DCSRM (see Section~\ref{sec:subemb} for details).}\label{fig:ssdd_dcsrm}
% \vspace{-5mm}
\end{figure*}

Figure \ref{fig:ssdd_dcsrm} illustrates the semantic alignment between the semantic rating data and each dimension of the transformed sub-embedding within the corresponding semantic dimension.

\section{Anatomically defined brain regions}
\begin{figure*}[htbp]
\centering 
\setlength{\textfloatsep}{2pt}
\setlength{\intextsep}{2pt}
\setlength{\abovecaptionskip}{2pt}
\includegraphics[width=0.9\textwidth]{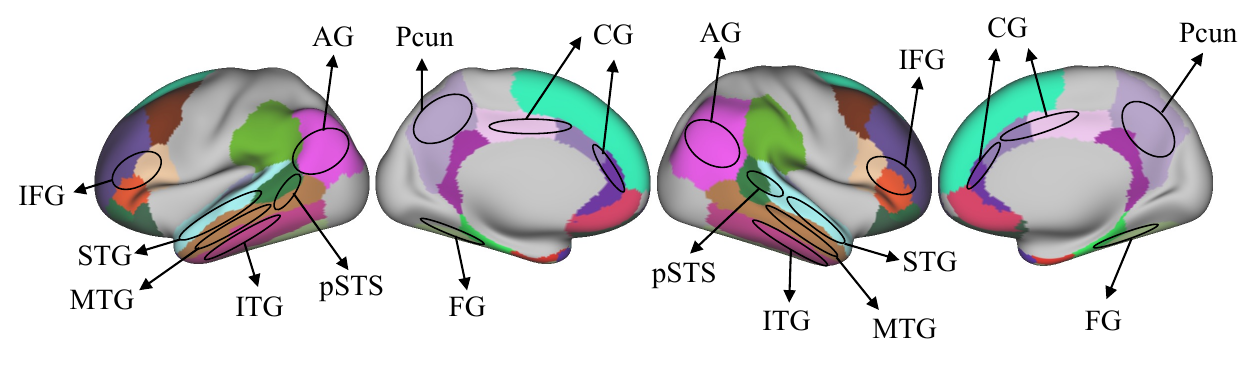}
\caption{\textbf{Anatomically defined brain regions involved in language processing.} The labeled regions are widely implicated in language comprehension and production, including the inferior frontal gyrus (IFG), superior temporal gyrus (STG), posterior superior temporal sulcus (pSTS), middle temporal gyrus (MTG), inferior temporal gyrus (ITG), precuneus (Pcun), cingulate gyrus (CG), fusiform gyrus (FG), and angular gyrus (AG).}
\label{fig:BNA}
% \vspace{-5mm}
\end{figure*}

Figure~\ref{fig:BNA} shows the anatomically defined brain regions that are commonly associated with language processing.

\section{Voxel-wise encoding results for SSDD and transformed sub-embeddings}
\begin{figure*}[htbp]
\centering 
\includegraphics[width=0.9\textwidth]{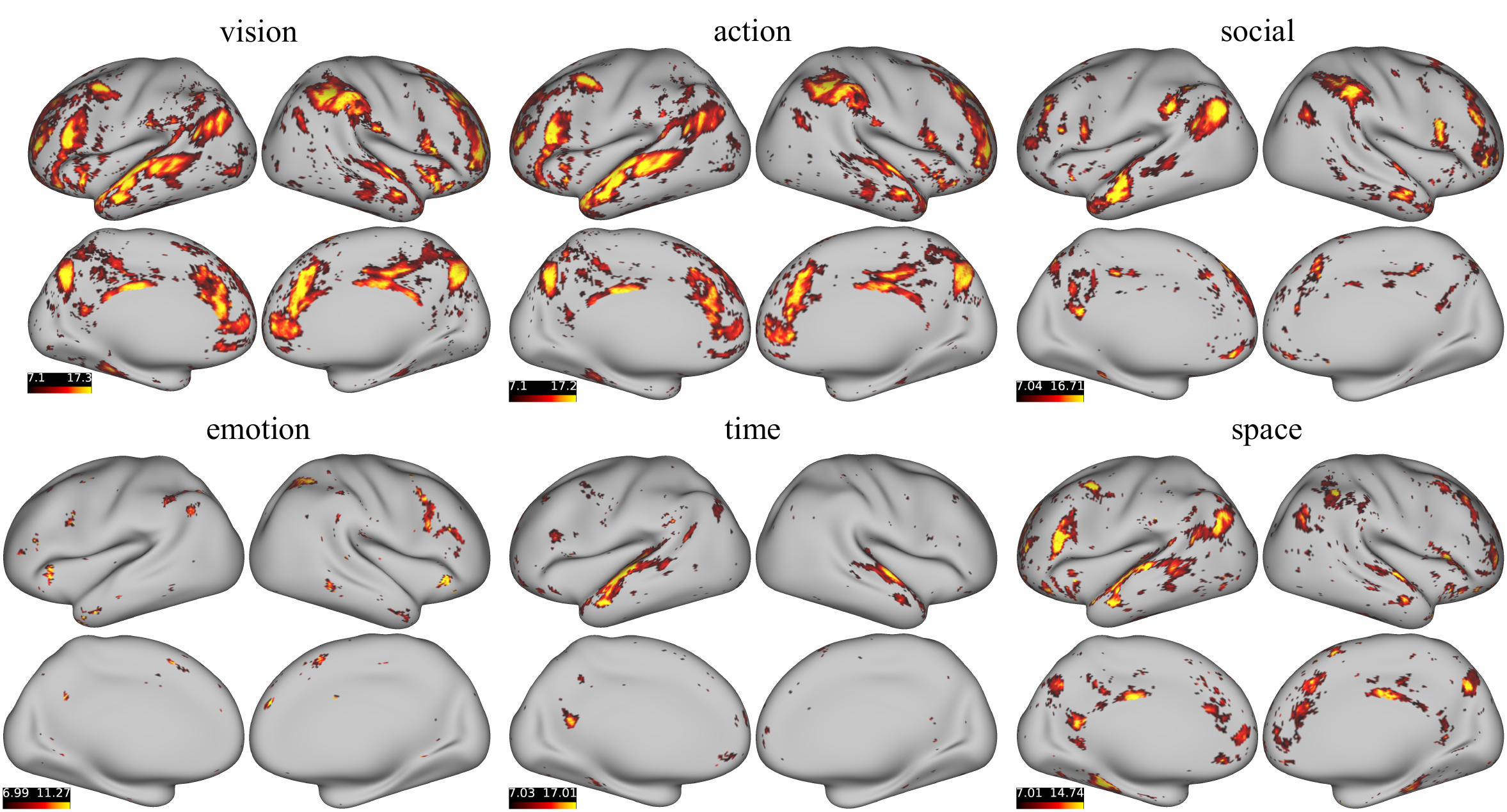}
\caption{\textbf{Group-level voxel-wise encoding results for SSDD.} Model performance was evaluated by computing the Pearson correlation between predicted and observed BOLD responses at each voxel. Color-highlighted regions indicate areas where BOLD responses predicted from SSDD rating data significantly exceeded a random baseline (one-tailed t-test, $-\log(p)$, $p < 0.001$).}
\label{fig:ssdd_encoding}
% \vspace{-5mm}
\end{figure*}

\begin{figure*}[htbp]
\centering 
\includegraphics[width=0.9\textwidth]{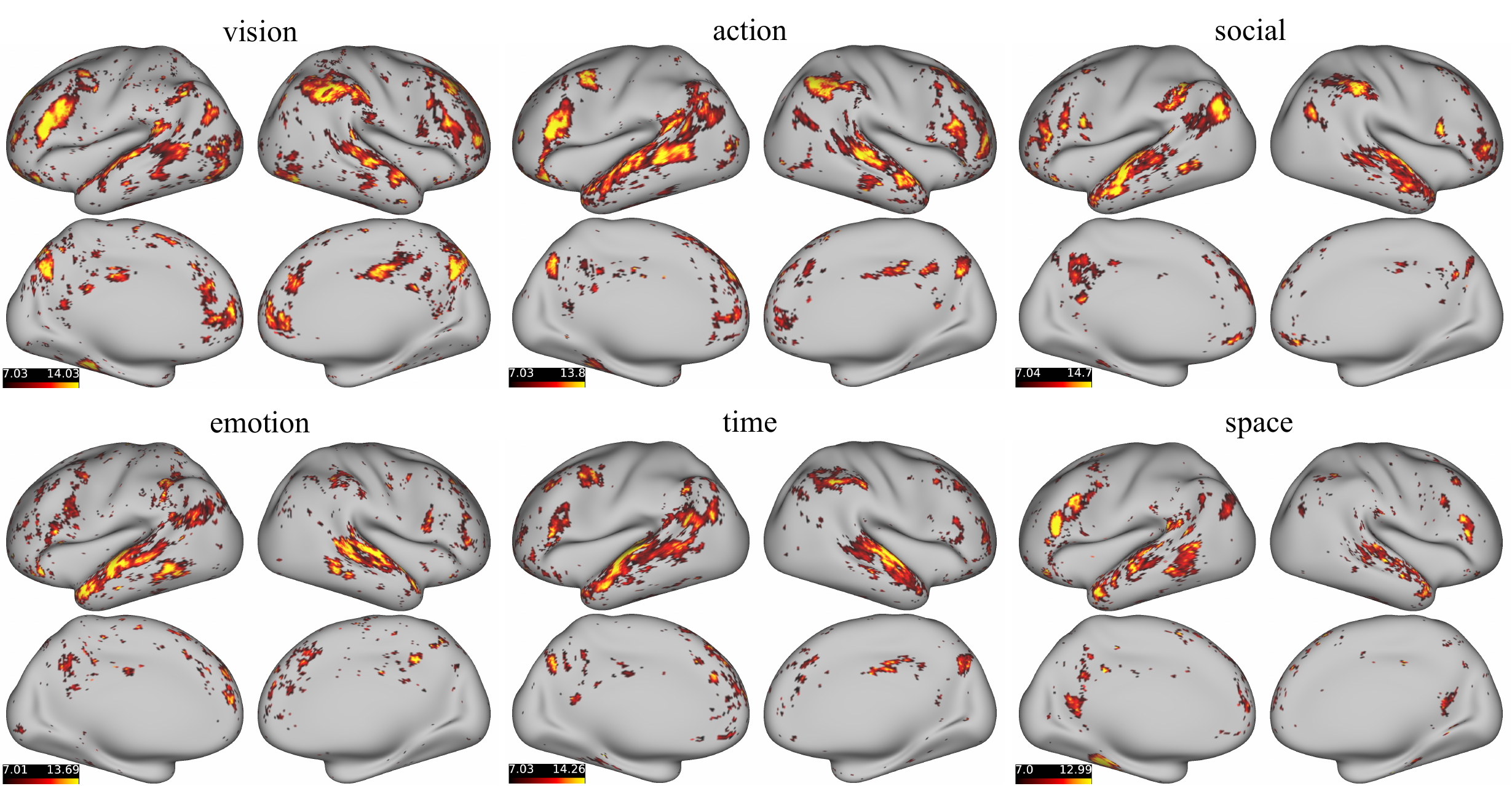}
\caption{\textbf{Group-level voxel-wise encoding results for transformed sub-embeddings.} Model performance was evaluated by computing the Pearson correlation between predicted and observed BOLD responses at each voxel. Color-highlighted regions indicate areas where BOLD responses predicted from DCSRM transformed sub-embeddings significantly exceeded a random baseline (one-tailed t-test, $-\log(p)$, $p < 0.001$). Transformed sub-embeddings refer to the outputs obtained by applying PCA individually to each semantic-specific sub-embedding produced by DCSRM (see Section~\ref{sec:subemb} for details).}
\label{fig:pca_encoding}
% \vspace{-5mm}
\end{figure*}

Figure~\ref{fig:ssdd_encoding} and Figure~\ref{fig:pca_encoding} show the group-level voxel-wise encoding performance for different semantic dimensions, using human rating data from SSDD and transformed sub-embeddings derived from DCSRM, respectively.

\section{Licenses of scientific artifacts}
\begin{table*}[htbp]

 \setlength{\tabcolsep}{2mm}
 \centering
 \small
 \renewcommand\arraystretch{1.25}
 \begin{center}
  \begin{tabular}{ll}
   \toprule[1.2pt]  
               \multicolumn{1}{l}{\textbf{Name}} & \multicolumn{1}{l}{\textbf{License}} \\
                \midrule[0.8pt]
                Transformers      & Apache 2.0 license     \\
                Connectome Workbench      &  GNU General Public license     \\
                NiBabel      & MIT license      \\
                Matplotlib      & PSF license      \\
                LLaMA3-8b      & Apache 2.0 license      \\
                Alpaca2-7b      & Apache 2.0 license      \\
                Alpaca2-1.3b      & Apache 2.0 license  \\
                LLaMA2-7b      & Apache 2.0 license      \\
                LLaMA2-1.3b      & Apache 2.0 license  \\
                MacBERT-large      & Apache 2.0 license  \\
                SMN4Lang     & CC BY-SA 4.0 and CC0 license  \\
                SSDD      & CC BY-SA 4.0 license  \\
   \bottomrule[1.2pt]
  \end{tabular}
 \end{center}
 \vspace{-2mm}
    \caption{\label{tab:license} Licenses of scientific artifacts involved in this work.}
 \vspace{-5mm}
\end{table*}
We follow and report the licenses of scientific artifacts involved in Table \ref{tab:license}.

\end{document}